\documentclass[conference]{IEEEtran}

\usepackage[utf8]{inputenc}
\usepackage[T1]{fontenc}
\usepackage{graphicx}
\graphicspath{{imagini/}}
\usepackage{amsmath,amssymb}
\usepackage{booktabs}
\usepackage{array}
\usepackage[hyphens]{url}
\usepackage{xcolor}
\usepackage{tabularx}
\usepackage{subcaption}
\usepackage{enumitem}
\usepackage[hidelinks]{hyperref}

\newcolumntype{L}{>{\raggedright\arraybackslash}X}
\newcommand{\yes}{\checkmark}
\newcommand{\no}{--}
\newcommand{\pit}{$\sim$}
\newcommand{\prt}{$\sim$}

\begin{document}

\title{TRISTAR: Triple-Signal Stair Recognition and\\
Vision-Only Indoor Navigation for\\
Search-and-Rescue Micro-UAVs}

\author{\IEEEauthorblockN{Octavian Gîngu\textsuperscript{*} and Stelian Spînu\textsuperscript{\dag}}
\IEEEauthorblockA{Military Technical Academy ``Ferdinand~I'', Bucharest, Romania\\
\textsuperscript{*}octavian.gingu@mta.ro \qquad \textsuperscript{\dag}stelian.spinu@mta.ro}
\thanks{A preliminary version of this work was presented at the CERC~2026
International Student Conference, Military Technical Academy, Bucharest, Romania.}}

\maketitle

\begin{abstract}
Indoor search-and-rescue (SAR) operations often require rapid situational
awareness where GNSS signals are unavailable and human access is difficult or
hazardous. While most autonomous aerial systems rely on LiDAR, stereo vision, or
specialized depth cameras, such solutions increase both hardware complexity and
deployment costs. This paper presents a complete autonomous indoor navigation
framework for low-cost unmanned aerial vehicles based exclusively on monocular
vision. Implemented on a DJI Tello platform, the system combines monocular depth
estimation using Depth Anything V2 with classical computer vision and lightweight
deep learning models for scene understanding, victim detection, and hazard
recognition. The framework consists of two independent behaviors: (i)~corridor
exploration with automatic door detection, room entry, OCR-based room
identification, and victim inspection; and (ii)~autonomous stair ascent based on
TRISTAR (TRI-Signal STair Ascent Recognition), a novel triple-sensor fusion
method that integrates structural cues (Sobel filtering), texture analysis
(multi-scale Gabor filtering), and geometric depth from monocular depth
estimation. Evaluation used real indoor flights in a university building. Depth
calibration reduced relative depth error from 27.4\% to below 10\%, while the
door detection algorithm reached a precision of 0.93 and an F1-score of 0.91. A
dedicated ablation study shows that multi-sensor fusion significantly improves
stair-recognition robustness compared to individual sensing modalities, and a
failure-case analysis delineates the limits of monocular perception under
challenging lighting and reflective surfaces. The results demonstrate that
reliable indoor exploration and stair traversal are achievable on
resource-constrained platforms without specialized ranging hardware---a
practical, cost-effective solution for rapid SAR deployment.
\end{abstract}

\begin{IEEEkeywords}
autonomous drones, indoor navigation, search and rescue, monocular depth
estimation, stair detection, sensor fusion, computer vision, UAV autonomy.
\end{IEEEkeywords}

\section{Introduction}
\IEEEPARstart{I}{ndoor} search-and-rescue (SAR) is, alongside urban rescue, one
of the most demanding classes of first-response intervention. Unlike operations
conducted in open space, indoors the rescuers have no satellite positioning,
visibility is degraded by dust, smoke, or darkness, and the building layout is
seldom known in advance. Under such conditions, every additional minute spent
locating the people inside sharply increases the risk to victims and rescuers
alike. Public statistics confirm the scale of the problem: in Romania alone,
residential-building fires exceed 6{,}900 cases per year, complemented by
thousands of in-building emergency-medical interventions,\footnote{General
Inspectorate for Emergency Situations (IGSU), \emph{Evaluation of IGSU activity
in 2024}.} a significant fraction of which involve spaces with difficult access
or reduced visibility. At the same time, data from FEMA and NIST indicate that
the critical time window for rescuing a trapped victim is on the order of
minutes, while the average time for a specialized team to penetrate an unknown
building remains well above that value. This gap justifies the continued
investment in technologies that give teams an informational head start before
they physically enter.

Small unmanned aerial vehicles (micro-UAVs) have become an increasingly
attractive option for this class of missions, thanks to their ability to
rapidly inspect narrow spaces without exposing human personnel. Yet ordinary
commercial drones remain fundamentally dependent on a trained operator, and
their real indoor autonomy is limited: in the absence of GNSS, most industrial
solutions resort to additional sensors (LiDAR, stereo depth cameras, high-grade
IMUs) that substantially increase the cost, mass, and complexity of the system.
Current commercial offerings fall broadly into two categories. On one side are
industrial drones such as Skydio or the professional DJI series (Matrice, Mavic
3 Enterprise), which are capable but costly (from 4{,}000 to over 25{,}000~EUR)
and dependent on dedicated ranging hardware. On the other side are lightweight
platforms that expose a programmatic API but, lacking a depth sensor, restrict
their real autonomy to point tasks such as face tracking or reactive obstacle
avoidance. Neither category directly covers the use case addressed here: a
low-cost drone able to autonomously explore a corridor, enter a room, scan it,
and report---something that, to date, has required either a human operator or an
expensive sensor suite.

Recent advances in monocular vision reopen this problem from a
software-centric angle. Transformer-based models for monocular depth estimation
now yield reasonably accurate metric depth directly from an ordinary video
stream, without a stereo rig or a LiDAR unit. In particular, Depth Anything
V2~\cite{depthanythingv2}, trained on diverse indoor and outdoor data, produces
per-pixel depth at a latency compatible with a real-time control loop on a
commodity GPU. If monocular depth is good enough to estimate the distance to a
wall, then it should also be good enough to let a drone navigate a corridor
autonomously and identify an open door. Building on this observation, our work
extracts as much navigation- and reporting-relevant information as possible from
a single video feed, instead of adding hardware to the drone. The solution is
built on a DJI Tello---accessible (list price under 150~EUR) and, although not
designed for complex autonomous flight, exposing an SDK sufficient to issue
remote-control commands at useful rates (10--12~Hz).

Beyond navigation proper, a realistic search mission entails complementary
tasks: identifying the rooms the drone enters (by reading the door number),
detecting people and estimating their condition (to dispatch the appropriate
medical team), and flagging visible hazards (fire, smoke). Dedicated neural
models already exist for each of these; the real challenge is not the models
themselves but their coherent integration into a system that operates without
continuous human intervention for the entire mission. The system is organized as
a local server with a web interface, letting the operator trigger and monitor
the mission in real time without directly piloting the drone. It supports two
distinct autonomous behaviors---corridor navigation and stair ascent---each
implemented as a separate module, individually validatable and triggered from the
operator interface according to the mission scenario.

\textbf{Contributions.} The main contributions of this paper are summarized as
follows.
\begin{enumerate}[leftmargin=*]
\item A \textbf{complete autonomous indoor-navigation system for low-cost drones
based exclusively on monocular perception}, with no specialized depth sensors,
integrating depth estimation, autonomous navigation, semantic scene analysis, and
automatic operational reporting into a single functional chain around one onboard
camera feed.
\item \textbf{TRISTAR} (TRI-Signal STair Ascent Recognition), a
\textbf{multi-sensor stair detector} that fuses structural, textural, and
geometric information obtained from Sobel filters, multi-scale Gabor filters, and
monocular depth estimation into a single composite confidence score.
\item An \textbf{appearance-free door-detection algorithm based on scene geometry
and a calibrated depth map}, requiring no dedicated trained detector, robust to
appearance traps, and functional even when the door is only partially in view.
\item A \textbf{dynamically switchable empirical depth-calibration scheme}
(room/corridor profiles) that reduces the relative monocular depth error from
27.4\% to below 10\%, enabling reliable control decisions without a ranging
sensor.
\item A \textbf{real-flight validation and ablation study} that quantifies the
contribution of each virtual sensor to the performance of the TRISTAR detector.
\item A \textbf{systematic failure-case analysis} exposing the specific
limitations of monocular perception in complex indoor environments (reflective
surfaces, glazing, low light).
\end{enumerate}

The implementation discussed in this paper is publicly available on GitHub at the
following repository:
\url{https://github.com/tavigingu/HawkOps-Indoor-Drone-Navigation-and-TRISTAR-Stair-Climbing}.

\section{Related Work}
Autonomous navigation with small drones has advanced rapidly, both through
commercial products and through academic contributions. We review the solutions
relevant to our use case---autonomous indoor exploration from a monocular video
feed---and identify the limitations that motivate our approach.

\subsection{Commercial drone platforms}
\textbf{Skydio}~\cite{skydio} (Skydio~2+ / X10) is among the best-known makers of
autonomous drones, capable of avoiding obstacles even in cluttered spaces thanks
to six 4K stereo cameras and an onboard NVIDIA Jetson processor. However, the
starting price of a Skydio~X10 exceeds 11{,}000~USD, the solution is entirely
black-box with no SDK to extend the navigation logic, and the supported missions
are limited to manufacturer-defined behaviors (tracking, patrol). \textbf{DJI}
professional models~\cite{dji} (Matrice~30/350~RTK) carry stereo cameras and ToF
sensors and support \emph{ActiveTrack}/\emph{Waypoint} modes outdoors, but an
equipped Matrice~30 exceeds 9{,}000~EUR, indoors without GNSS it falls back on a
Vision Positioning System that demands adequate lighting and floor texture, and
there is no native mode for entering rooms or reading door labels. \textbf{Modal
AI}~\cite{modalai} (Starling~2 with Voxl~2) targets developers with a ROS/PX4
stack and monocular SLAM, but a starter kit costs about 5{,}500~USD, requires
PX4/ROS expertise, and provides no autopilot for complex ``explore the corridor
and enter each room'' missions. Finally, the \textbf{DJI Tello} educational
drone ($\sim$80~g, under 150~EUR) exposes a UDP SDK abstracted by the
open-source \texttt{djitellopy}~\cite{djitellopy} library; it carries no
dedicated depth sensor---its only external signal is a 720p frontal video
stream---which makes it an attractive research platform precisely because any
behavioral improvement is a consequence of the software layer above the SDK, and
is therefore directly portable and reproducible.

\subsection{Monocular depth estimation}
Estimating depth from a single image is a classic problem accelerated by
transformer models trained on diverse datasets. \textbf{MiDaS}~\cite{midas}
popularized scale-invariant relative depth from a single image but produces only
relative depth. \textbf{ZoeDepth}~\cite{zoedepth} adds a metric regression head
(trained on NYUv2~\cite{nyuv2} and KITTI~\cite{kitti}) yet its GPU latency can be
prohibitive for real-time loops above 10--15~Hz. \textbf{Depth Anything
V2}~\cite{depthanythingv2}, adopted in this work, is pretrained on a large volume
of synthetic data and fine-tuned on real data; with a ViT-S backbone it runs
above 20~Hz at 360p on a consumer GPU and, crucially, exhibits markedly better
temporal stability on video than MiDaS or ZoeDepth---essential for stable door
and step detection. Its relative output is converted to metric distances through
an empirical calibration built on real scenes (Section~\ref{sec:calib}), which
proved more stable on the specific geometry of the targeted corridors and rooms
than a pretrained metric variant.

\subsection{Academic prototypes and structure detection}
A first group of works uses monocular depth (MiDaS, Depth Anything) for reactive
obstacle avoidance on drones, but without high-level logic (room entry, label
reading, reporting). A related current is neural monocular SLAM
(DROID-SLAM~\cite{droidslam}, DPVO~\cite{dpvo}), which reconstructs a globally
consistent 3D map---a more ambitious goal than the frame-by-frame reactive
navigation we pursue, and one whose online computational cost remains high for a
10--12~Hz loop on a small commercial drone. Door detection has been addressed
with RGB-D data: Arduengo et al.~\cite{arduengo} operate doors with a mobile
robot, estimating the parameters needed for autonomous opening, while Ramôa et
al.~\cite{ramoa} build an RGB-D dataset and classify door state on a low-power
device; both rely either on dedicated depth sensors or on models trained on the
visual appearance of the door. YOLO-based door detectors~\cite{park,lin} work
reasonably on static images but are unstable on moving video and are
fundamentally unsuited to our central case---an \emph{open} door the drone flies
through: as the drone approaches, the door frame, threshold, and handle leave the
field of view, leaving an appearance-trained detector with no cues. In essence,
our system does not detect a door as an object, but a structural opening in an
otherwise flat wall---a more distant region surrounded by nearby surfaces---a
configuration described far more naturally through the depth map than through an
appearance model. Autonomous stair ascent by drones is addressed only
occasionally, typically with LiDAR, stereo, or fiducial markers, with
purely monocular solutions being rare and limited to simple demonstrators.
Integrated indoor SAR systems appear at events such as the DARPA Subterranean
Challenge~\cite{darpasubt}, but these use industrial-grade drones with LiDAR and
additional sensors, at a total platform cost far above that of an accessible,
low-cost system.

For the detection of regular structures (stair steps), our work leans on
established classical techniques. The Sobel operator yields a directional
gradient whose vertical modulus $|G_y|$ responds strongly to each horizontal step
edge, forming a 1D profile of regular peaks. Gabor filters~\cite{gabor1946},
orientation-selective at $\theta=90^\circ$ and tuned to several wavelengths,
respond much more strongly to horizontal lines (steps) than to vertical ones
(doors, railings). The Hough transform would be a natural alternative but tends
to produce many false positives on partially occluded steps or under strong
perspective; we therefore prefer the Sobel/Gabor combination, complemented by a
third virtual sensor derived from the depth map.

\subsection{Positioning}
Table~\ref{tab:comparison} summarizes the functional comparison against the most
representative commercial and academic solutions. To the best of our knowledge,
none of the studied commercial systems covers, without additional programming
effort, all four critical capabilities---corridor navigation with automatic room
entry, autonomous stair ascent, OCR label reading, and AI-based medical
assessment of detected people---which the proposed system delivers starting from
a drone costing under 150~EUR, with spatial perception that is exclusively
monocular. The design trade-offs implied by these choices (no global map, strict
separation of the two behaviors) are stated explicitly and analyzed in
Section~\ref{sec:conclusion}.

\begin{table}[t]
\centering
\caption{Functional comparison of existing solutions and the proposed system.}
\label{tab:comparison}
\footnotesize
\setlength{\tabcolsep}{3pt}
\begin{tabular}{@{}lccccc@{}}
\toprule
\textbf{Capability} & \rotatebox{60}{Skydio X10} & \rotatebox{60}{DJI Matrice 30}
& \rotatebox{60}{Modal AI Starling} & \rotatebox{60}{Tello + lib.}
& \rotatebox{60}{\textbf{Proposed}} \\
\midrule
HW cost (EUR)                & $>$11k & $>$9k & \prt5.5k & $<$150 & $<$150 \\
Hardware depth sensor        & \yes   & \yes  & \yes      & \no    & \no    \\
Open SDK                     & \pit   & \yes  & \yes      & \yes   & \yes   \\
Indoor nav.\ (no GNSS)       & \yes   & \pit  & \yes      & \no    & \textbf{\yes} \\
Autonomous room entry        & \no    & \no   & code      & code   & \textbf{\yes} \\
OCR label reading            & \no    & \no   & code      & code   & \textbf{\yes} \\
Autonomous stair ascent      & \no    & \no   & code      & code   & \textbf{\yes} \\
Person + AI triage           & \pit   & \pit  & code      & code   & \textbf{\yes} \\
Post-mission report          & \no    & \pit  & code      & code   & \textbf{\yes} \\
\bottomrule
\end{tabular}
\vspace{2pt}
\footnotesize\raggedright \yes: supported; \pit: partial; \no: absent;
``code'': requires full user implementation.
\end{table}

\section{Methods}
\label{sec:methods}

\subsection{System architecture}
\begin{figure}[t]
\centering
\includegraphics[width=\columnwidth]{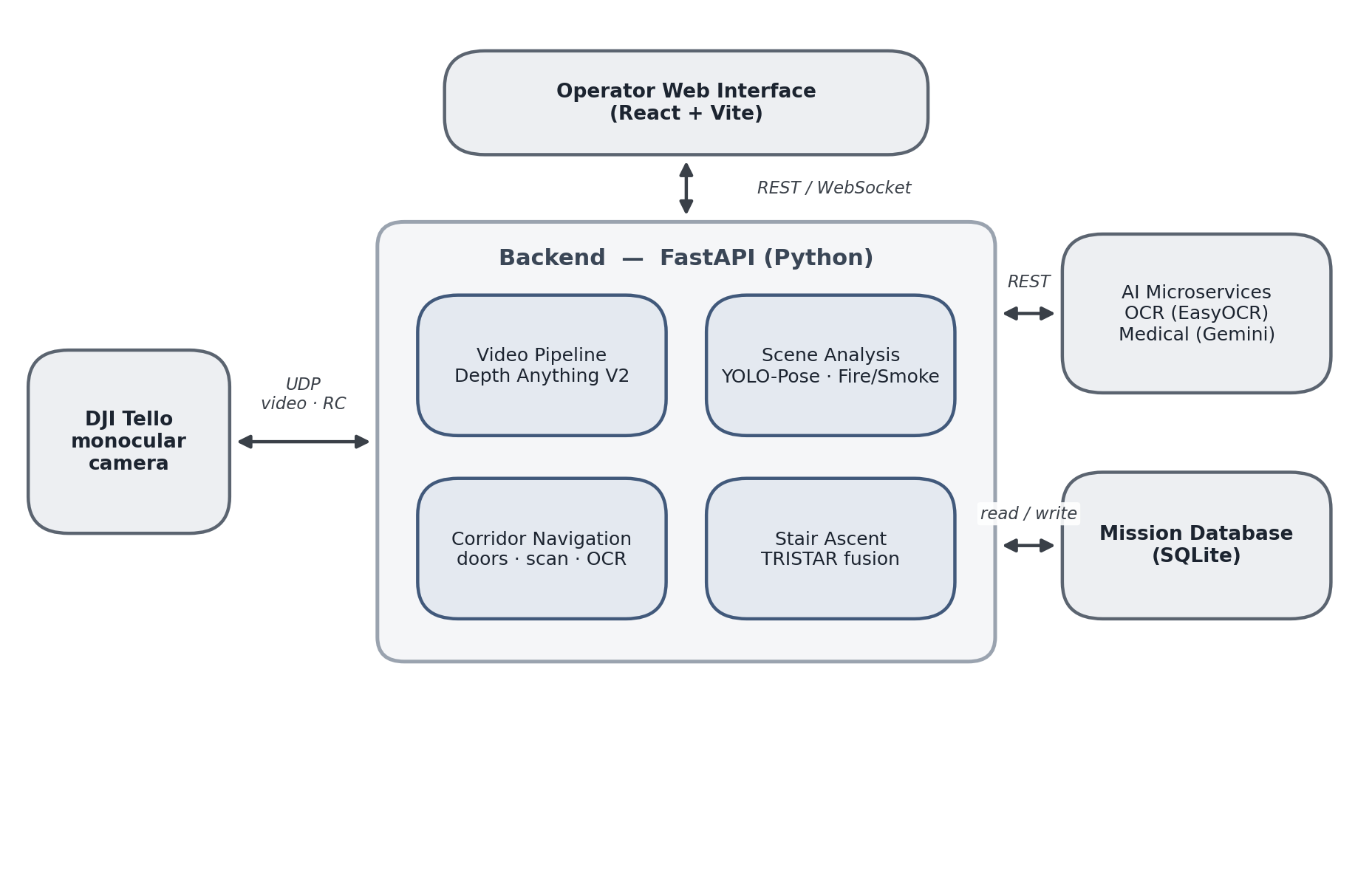}
\caption{Overall system architecture. The FastAPI backend runs all perception
locally (monocular depth, corridor navigation, TRISTAR stair ascent, scene
analysis) and communicates with the operator web interface, the DJI Tello, a
mission database, and two optional AI microservices (OCR, medical) that can be
disabled when there is no connectivity.}
\label{fig:architecture}
\end{figure}

The system is organized as five components communicating over REST and WebSocket
on the local network: a \textbf{backend} (FastAPI~\cite{fastapi} plus vision
models), a \textbf{web frontend} (React), \textbf{two external microservices}
(OCR and medical AI) called on demand, the \textbf{DJI Tello} drone (connected by
UDP), and a \textbf{database} that persists missions and results
(Fig.~\ref{fig:architecture}). The backend is the core: \texttt{server.py}
mounts the endpoints and manages the drone lifecycle; \texttt{tello\_controller}
wraps \texttt{djitellopy} behind a stable API; \texttt{stream\_pipeline} captures
frames and runs Depth Anything V2; the \texttt{navigation} package holds the
corridor logic; and \texttt{navigation/stair\_climber} holds the fully
independent stair module. The pipeline runs three neural models concurrently on
GPU---Depth Anything V2 (ViT-S) for depth, YOLO11n-Pose~\cite{yolo11} for people
and 17 body keypoints, and YOLOv8~\cite{yolov8} for fire/smoke---alongside video
decoding and the autopilot, in distinct Python threads. Two microservices with
heavy dependencies are isolated: OCR (EasyOCR~\cite{easyocr}) and medical AI
(Gemini~2.5 Flash~\cite{gemini}); both are optional for navigation and can be
disabled when offline. The total size of the locally loaded models is
$\approx$200~MB.

Several challenges shaped these design decisions. The Tello control channel
accepts \texttt{send\_rc\_control(lr,fb,ud,yaw)} commands only at 10--15~Hz over
UDP; the H.264 video stream introduces 80--150~ms of decoding latency; and, on a
moving drone, the metric depth of Depth Anything V2 can oscillate by
$\pm0.2$~m between consecutive frames, especially on homogeneous surfaces. To
avoid unstable state switching, all derived metrics are smoothed by an
exponential moving average (EMA); phase transitions additionally apply a
symmetric \emph{deadband} around the target value and accept a new state only
after the EMA metric stays within the band for several consecutive cycles
(\texttt{stable\_hits}$=3$), which eliminates the pendular transitions caused by
frame-to-frame fluctuations. Because a control error has physical consequences,
each phase has a timeout, a low-battery forced-landing condition, an operator
abort button, and guard-clause checks at the start of every control loop.

\subsection{Video pipeline and depth calibration}
\label{sec:calib}
A dedicated thread reads the last decoded frame from the SDK at a target of
30~FPS and places it into a bounded, drop-oldest queue: in real time, only the
latest frame matters. Depth Anything V2 (ViT-S) is loaded from a local
checkpoint and, in the configuration used here, runs the \emph{relative} variant;
the relative map is converted to metric distances by an empirical calibration
layer, which grants full control over the correspondence between the model's raw
value and the true distance and adapts it to the targeted geometry.

The calibration is a lookup table mapping the raw value (blue-channel luminance
of the colored map, $0$--$255$) to the real distance measured with a laser
rangefinder, interpolated linearly. It is organized into dynamically switchable
\emph{profiles}: \texttt{room} (a furnished room) and \texttt{corridor} (a long
hallway with plain walls). Profiles are switched mid-mission through a REST
endpoint---the corridor autopilot switches to \texttt{room} after crossing a
door and back to \texttt{corridor} on exit. From the calibrated map, per-frame
global metrics feed both the autopilot and the interface, the most important
being the global coefficient of variation (CoV) and the ``blue ratio'':
\begin{equation}
\label{eq:cov}
\mathrm{CoV} = \frac{\sigma(D)}{\mu(D)}, \qquad
r_{\text{blue}} = \frac{\bigl|\{\,p\in D \mid d_{\min}\le D(p)\le d_{\max}\}\bigr|}{|D|},
\end{equation}
where $D$ is the set of per-pixel metric depths. Steps produce a large CoV; a
flat wall a small one. The blue ratio acts as a proxy for how ``far'' the scene
extends and is central to the door-distance control loop.

\subsection{Corridor navigation}
\label{sec:corridor}
The corridor module explores a hallway and enters each open room in turn. The
drone's primary motion is a \emph{lateral crab to the right}: it keeps a constant
yaw and sweeps the corridor searching for an open door, then runs a sequence of
phases---centering, distance control, entry, scan, and exit---before resuming the
crab to find the next door.

\subsubsection{Door detection}
\begin{figure}[t]
\centering
\includegraphics[width=\columnwidth]{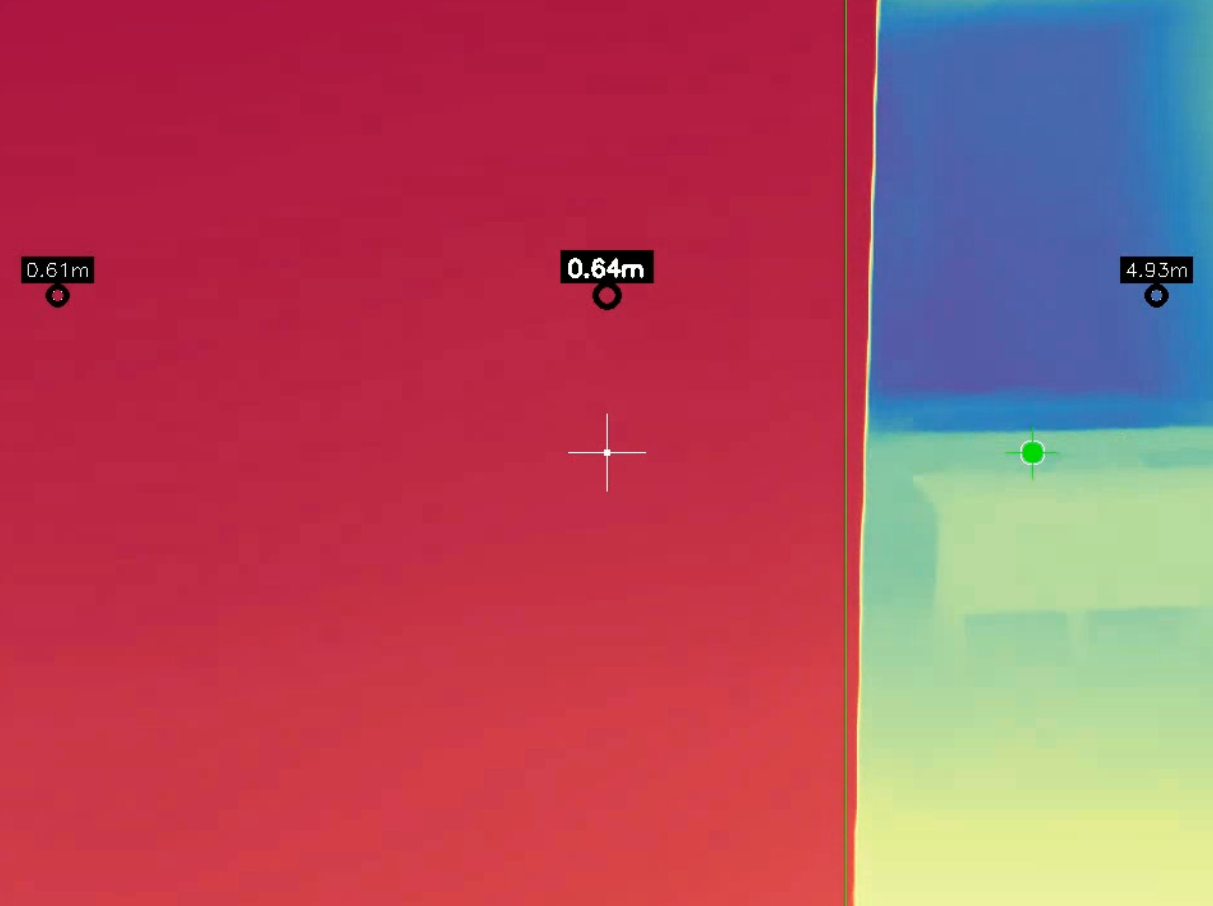}
\caption{Door detection on the depth map. The open door is isolated as a distant
region flanked by sharp vertical depth edges corresponding to the door casing;
appearance cues (handle, frame) are not required.}
\label{fig:door}
\end{figure}

Door detection uses \emph{only} the depth map, a deliberate choice justified by
the intrinsically spatial nature of an open door: it appears as a region
significantly more distant than the surrounding walls. A binary mask keeps pixels
between 1.7~m and 6.0~m; morphological \emph{close}/\emph{open}/\emph{dilate}
operations with a $7\times7$ structuring element consolidate contours; contours
are extracted and each candidate is filtered by three successive tests. The
\emph{shape} filter requires at least four vertices, a width/height ratio in
$[0.35,2.0]$, a height of at least 30\% of the frame, and a contour spanning from
the top toward the bottom. The \emph{content} filter requires fewer than 50\% of
the candidate's pixels to be closer than 1.5~m: an open door is an opening, not a
nearby surface. The \emph{vertical-edge} filter is the most important: on the
candidate's depth ROI, the horizontal gradient $|G_x|$ at the left and right
margins must be significantly stronger than in the center. Concretely, with
$Q_{0.9}$ the 90th percentile of $|G_x|$,
\begin{equation}
\label{eq:edge}
R = \frac{Q_{0.9}\bigl(|G_x|_{\text{margin}}\bigr)}
         {Q_{0.9}\bigl(|G_x|_{\text{center}}\bigr)} \ge 1.20 ,
\end{equation}
must hold (a single valid edge suffices, at a stricter $R\ge1.45$, when the door
touches the frame border). This exploits the fact that an open door is framed by
the door casing, which produces strong vertical edges in the depth map. Among the
survivors, the candidate maximizing $R\times\text{area}$ is selected
(Fig.~\ref{fig:door}).

\subsubsection{Centering, distance control and entry}
Once a door is detected, the drone aligns its frontal axis with the door axis
through a proportional lateral loop with a deadband,
$v_{lr}=\mathrm{clip}\!\bigl(K_{lr}(x_{\text{door}}-x_{\text{center}}),
-v_{\max},+v_{\max}\bigr)$, with $K_{lr}\approx0.10$ and a $\sim$25~px tolerance,
issued as short pulses with pauses to let the drone settle. Distance is then
regulated using \texttt{full\_frame\_blue\_ratio}, whose interpretation is
counter-intuitive: when the drone is too close, the door frame leaves the field
of view and the drone sees through the opening into the room, so most pixels
become distant---a \emph{high} value therefore means \emph{too close}, and the
drone must back off. The metric is filtered by an EMA,
\begin{equation}
\label{eq:ema}
r_{\text{EMA}}^{(t)} = \alpha\, r^{(t)} + (1-\alpha)\, r_{\text{EMA}}^{(t-1)} ,
\end{equation}
with $\alpha=0.18$ and a target $r^\star$ within a $\pm0.05$ deadband;
stabilization is declared after three consecutive in-band frames. Entry advances
a fixed 1.8~m using the drone's visual-odometry position command; the calibration
profile is then switched to \texttt{room}. Before entry, the room number is read:
the label plate is localized locally (Canny edges, aspect/area filtering), the
crop is Base64-encoded and sent to the OCR microservice, and the highest-scoring
text with a valid $2$--$4$-digit format is stored as \texttt{room\_label}.

\subsubsection{Scanning, inspection and exit}
Two scan modes are offered. In \emph{Simple} mode the drone measures the distance
to the front wall, advances about a third of it, performs a full $360^\circ$
rotation with brief pauses every $90^\circ$ for stabilization, and returns
symmetrically to the entry point. In \emph{Complex} mode it measures the front,
left, and right walls and then executes a five-segment perimeter crab covering a
half-perimeter of the room---the most exhaustive coverage, at the cost of time
and battery. Throughout the scan, the AI Analyzer runs in parallel: YOLO11n-Pose
detects people with 17 COCO~\cite{cocodataset} keypoints, ByteTrack~\cite{bytetrack}
maintains identities across frames, a purely geometric classifier assigns a
posture (\emph{standing}, \emph{sitting}, \emph{fallen}, \emph{unknown}), and for
each qualified person a cropped image is sent asynchronously to the medical
microservice, which returns a visible-state estimate (STABLE, INJURED, CRITICAL,
DECEASED, UNKNOWN). A second thread runs the fire/smoke detector, whose
detections are logged as report events. On exit, the drone turns
$205^\circ$ (empirical yaw-drift compensation), re-acquires the door by a crab
search, re-centers, re-stabilizes distance, and advances 1.8~m back into the
corridor with active lateral correction. Every visited room yields a JSON record
(OCR number, detected people with posture/medical state, hazards, timestamps)
persisted to a local SQLite database and rendered directly in the web interface.

\subsection{Autonomous stair ascent: the TRISTAR detector}
\label{sec:tristar}
\begin{figure*}[t]
\centering
\begin{subfigure}[t]{0.32\textwidth}\centering
\includegraphics[width=\linewidth]{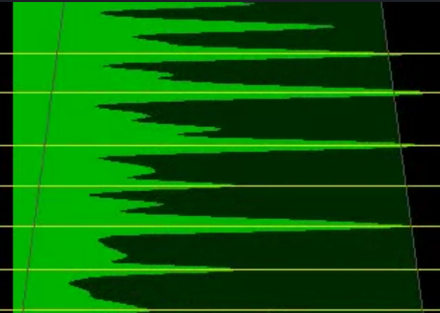}\caption{Sobel --- stairs}\end{subfigure}\hfill
\begin{subfigure}[t]{0.32\textwidth}\centering
\includegraphics[width=\linewidth]{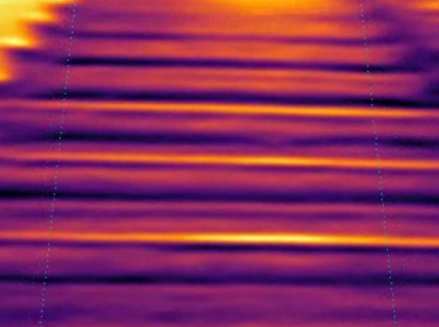}\caption{Gabor --- stairs}\end{subfigure}\hfill
\begin{subfigure}[t]{0.32\textwidth}\centering
\includegraphics[width=\linewidth]{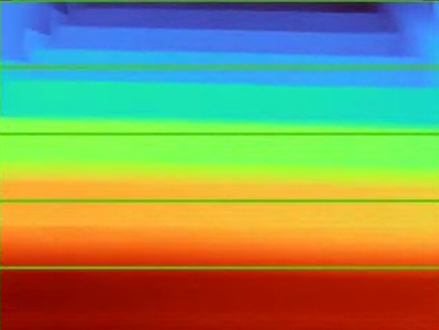}\caption{Depth (DA2) --- stairs}\end{subfigure}
\\[0.4em]
\begin{subfigure}[t]{0.32\textwidth}\centering
\includegraphics[width=\linewidth]{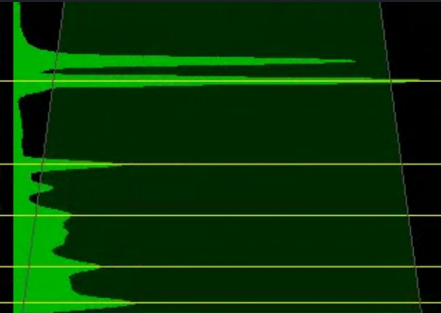}\caption{Sobel --- flat}\end{subfigure}\hfill
\begin{subfigure}[t]{0.32\textwidth}\centering
\includegraphics[width=\linewidth]{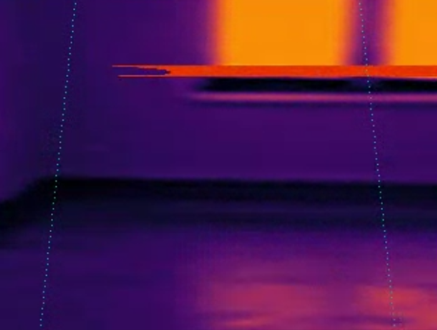}\caption{Gabor --- flat}\end{subfigure}\hfill
\begin{subfigure}[t]{0.32\textwidth}\centering
\includegraphics[width=\linewidth]{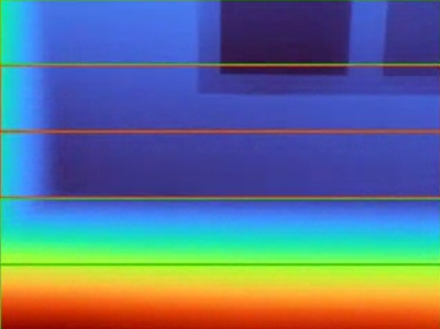}\caption{Depth (DA2) --- flat}\end{subfigure}
\caption{The three virtual sensors of TRISTAR on a staircase (top) and on a flat
surface (bottom). On stairs, Sobel yields regular horizontal peaks, Gabor yields
consistent edge-to-edge lines, and the depth map is monotonic across horizontal
bands; on a flat surface all three responses collapse.}
\label{fig:sensors}
\end{figure*}

The stair module, fully independent of the corridor module, drives the drone from
the base of an interior staircase to the upper landing---a flat surface large
enough to stabilize, with no visible steps. Its core is \textbf{TRISTAR}
(TRI-Signal STair Ascent Recognition), which fuses three independent virtual
sensors into a per-frame composite confidence score. Ascent starts only if the
composite score exceeds $0.52$ with corroborating validation from all three
sensors. Figs.~\ref{fig:sensors} illustrate the three signals on a staircase
versus a flat surface.

\subsubsection{Sensor 1 --- horizontal Sobel}
On a $320\times240$ analysis image, the vertical gradient modulus $|G_y|$
responds strongly to each horizontal step edge. A trapezoidal mask excludes
railings and side walls; the 1D vertical profile (row-wise mean) is Gaussian-%
smoothed, and peaks are detected by non-maximum suppression. A peak counts as a
valid step only if its line covers at least 55\% of the frame width. The score
combines the uniformity $U$ of inter-peak spacing (low CoV), the vertical
coverage $C$, and the peak density $D$, penalizing scenes dominated by vertical
gradients (likely doors or walls):
\begin{equation}
\label{eq:sobel}
S_{\text{sobel}} = 0.40\,U + 0.35\,C + 0.25\,D .
\end{equation}

\subsubsection{Sensor 2 --- multi-scale Gabor}
Five Gabor kernels at $\theta=90^\circ$ with wavelengths
$\lambda\in\{14,22,32,45,65\}$~px cover steps from $\sim$0.5~m to $\sim$4~m. When
depth is available, the wavelength is selected adaptively,
\begin{equation}
\label{eq:lambda}
\lambda^\star(d) = \frac{0.20}{d}\, f_y ,
\end{equation}
where 0.20~m is a standard step height, $d$ the estimated metric distance, and
$f_y$ the vertical focal length. The key validation criterion is
\emph{edge-to-edge} coverage: every strongly responding line must span from the
left to the right frame margin (with perspective-adjusted tolerance). Lines that
fail are marked ``red'', and
\begin{equation}
\label{eq:gabor}
S_{\text{gabor}} = C_{\text{edge-to-edge}}\,\bigl(1 - r_{\text{red}}^{\,0.35}\bigr) .
\end{equation}

\subsubsection{Sensor 3 --- monocular depth (DA2)}
Two depth properties are evaluated. The \emph{global CoV} is large on steps
($\mathrm{CoV}>0.28$) and small on flat surfaces ($\mathrm{CoV}<0.12$). The
\emph{band monotonicity} splits the image into five horizontal bands: on a
staircase viewed from below, the median depth must increase monotonically from
the bottom band (near step) to the top band (far step). A single order inversion
collapses the factor $M$ to $0.05$, two inversions to $0$, which rejects flat
walls or floors whose depth is large but non-monotonic. The score is
\begin{equation}
\label{eq:da2}
S_{\text{da2}} = \min\bigl(1,\ \mathrm{normalize}(\mathrm{CoV})\bigr)\cdot M .
\end{equation}

\subsubsection{Composite score and ascent loop}
The composite confidence is the equally weighted mean,
\begin{equation}
\label{eq:conf}
\mathrm{conf}_{\text{stairs}} = \tfrac{1}{3}\bigl(S_{\text{sobel}} + S_{\text{gabor}} + S_{\text{da2}}\bigr) ,
\end{equation}
with an ascent threshold of $0.52$. Once above threshold, the drone enters the
ascent loop at a fixed 12~Hz, issuing three simultaneous RC commands per tick.
The \emph{forward} speed brakes proportionally with the median distance to the
steps in the central region,
\begin{equation}
\label{eq:fwd}
v_{\text{fb}}(d) =
\begin{cases}
0 & d < d_{\min},\\[2pt]
v_{\max}\dfrac{d-d_{\min}}{d_{\text{tgt}}-d_{\min}} & d_{\min}\le d < d_{\text{tgt}},\\[6pt]
v_{\max} & d\ge d_{\text{tgt}},
\end{cases}
\end{equation}
with $d_{\min}=0.7$~m, $d_{\text{tgt}}=1.2$~m, $v_{\max}=30$~cm/s. The
\emph{vertical} speed is proportional to the local slope estimated from the depth
map by comparing the median depth of the upper band ($d_{\text{top}}$) and the
lower band ($d_{\text{bot}}$),
\begin{equation}
\label{eq:slope}
\theta = \arctan\!\bigl(\Delta d / y_{\text{span}}\bigr), \quad
v_{\text{ud}} = v_{\text{fb}}\,\tan\theta\, f_{\text{dist}} ,
\end{equation}
where $\Delta d = d_{\text{top}}-d_{\text{bot}}$ and $y_{\text{span}}$ is the
metric vertical extent of the frame. \emph{Lateral} correction is derived from
the left/right asymmetry of the Sobel step gaps. Ascent stops when all three
sensors report ``flat surface'' for five consecutive frames (Sobel flat score
$\ge0.40$, no edge-to-edge Gabor lines, $\mathrm{CoV}<0.12$ with lost
monotonicity); a short \emph{pre-landing dash} then moves the drone clear of the
last step edge. Safety layers include a 50~s per-flight timeout, a 20\%
battery floor, an operator abort, and continuous recording of every frame
(raw + debug overlay) for post-flight analysis.

\section{Experimental Results}
Given the physical nature of the application, validation was organized on three
levels: unit and integration tests on static datasets; reproducible flights with
full video and RC logging, analyzed post-flight; and real end-to-end flights
inside a university building. The backend ran on a laptop with an Intel Core~i7,
an NVIDIA RTX~3060 Mobile (6~GB VRAM), 16~GB RAM, Python~3.11, PyTorch~2.4, and
CUDA~12.1. All flights used a single standard DJI Tello (720p camera, $\sim$80~g,
$\sim$13~min per battery). Testing took place in a mock-up room (for calibration)
and in a university corridor ($\sim$2.4~m wide, $30$+~m long, doors on both sides)
and stairwell (standard steps, $\sim$0.17~m rise, $\sim$0.28~m run).

\subsection{Video pipeline and depth accuracy}
\begin{figure}[t]
\centering
\begin{subfigure}[t]{0.48\columnwidth}\centering
\includegraphics[width=\linewidth]{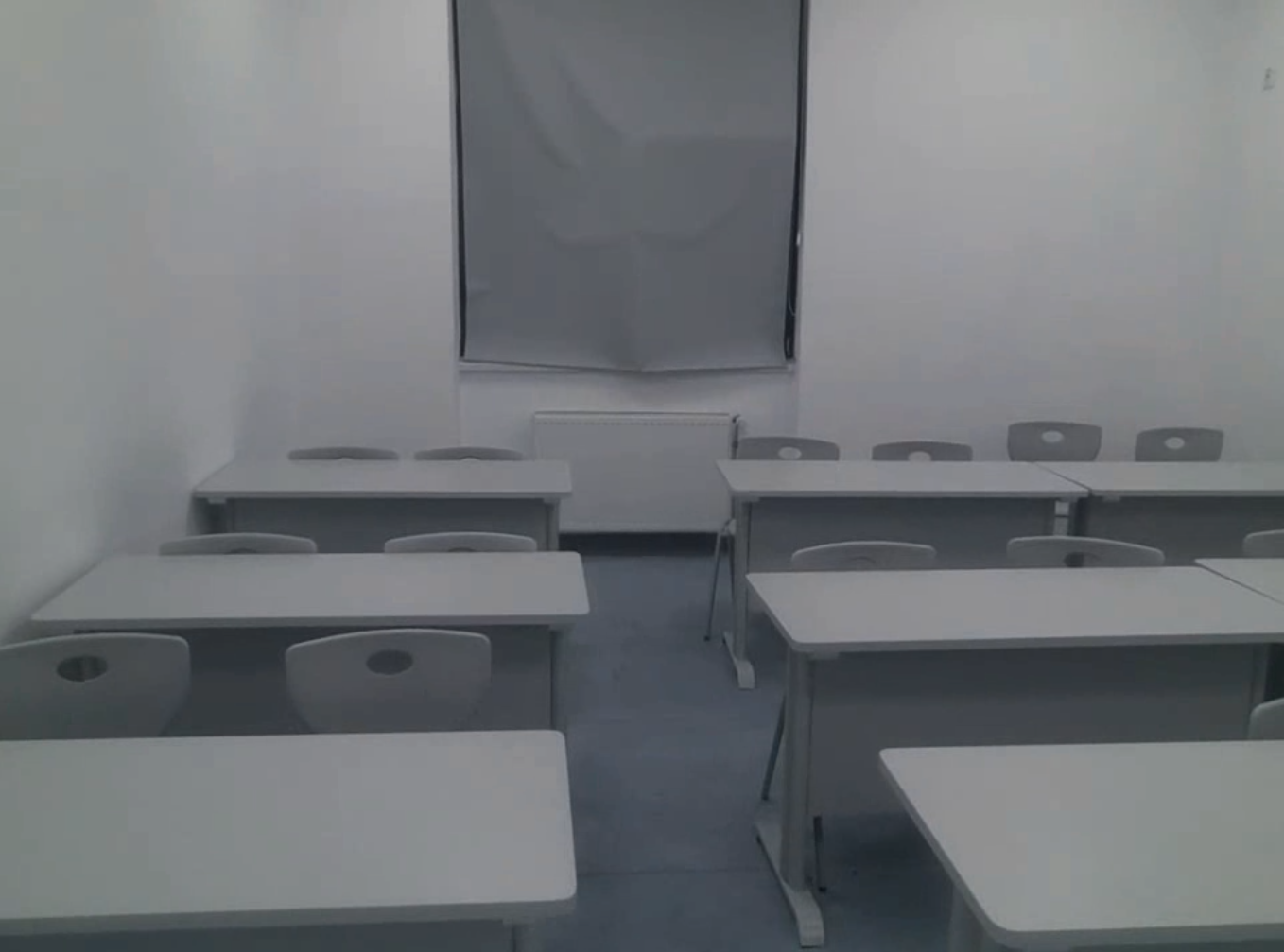}\caption{Raw RGB frame}\end{subfigure}\hfill
\begin{subfigure}[t]{0.48\columnwidth}\centering
\includegraphics[width=\linewidth]{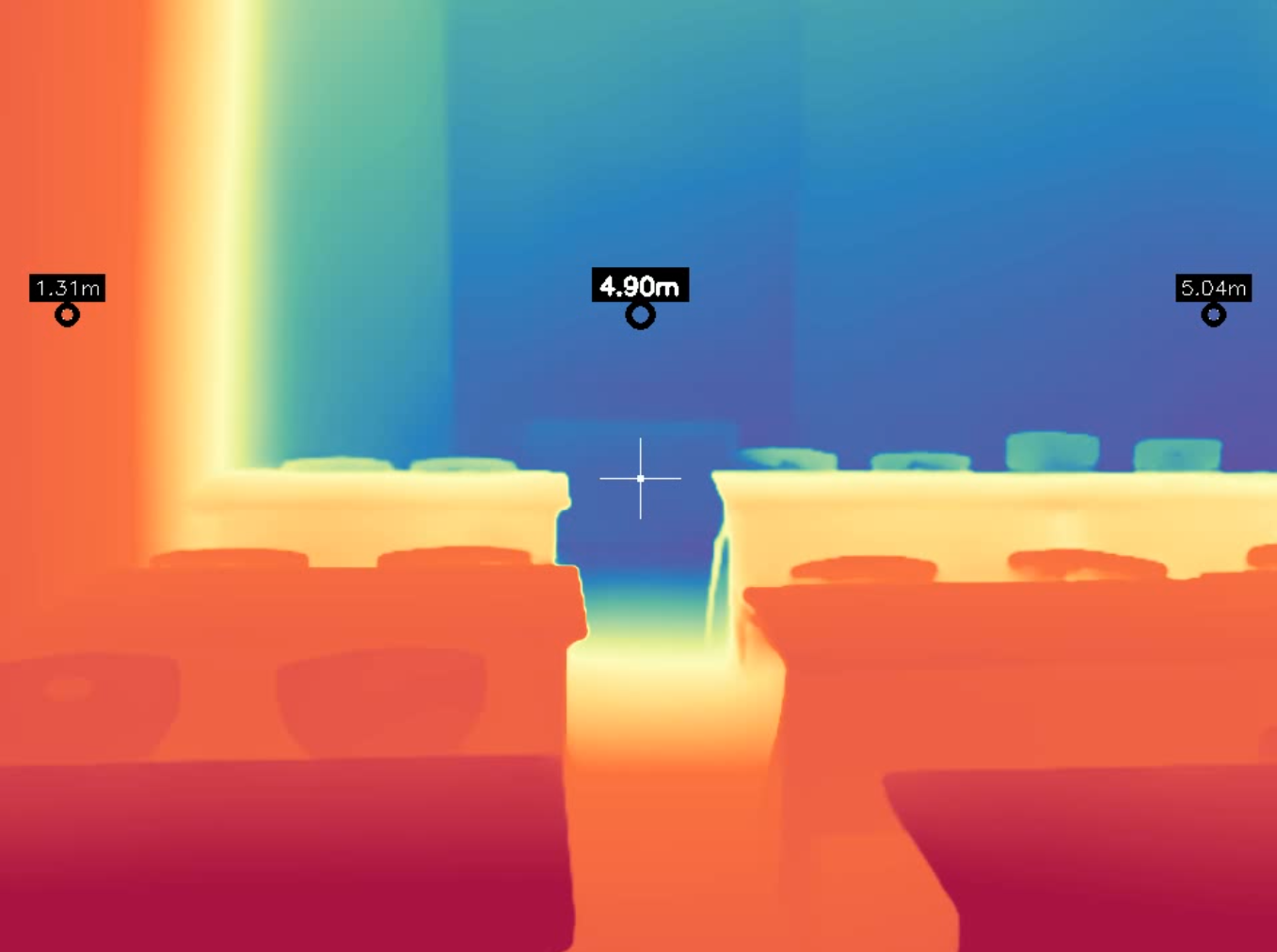}\caption{Calibrated depth map}\end{subfigure}
\caption{A raw frame (left) and its corresponding calibrated depth map (right),
the source of all derived navigation metrics.}
\label{fig:depthmap}
\end{figure}

The pipeline meets its non-functional targets: $\sim$30~FPS on the raw stream and
$\sim$10--12~Hz end-to-end (frame~$\to$~RC decision), with Depth Anything V2
(ViT-S) inferring at $\sim$45~ms per frame on GPU and an end-to-end latency of
$\sim$85--100~ms (well below the 300~ms budget). The depth calibration was
validated on 24 reference images across three contexts, with 5--8 manually
selected reference points per image measured by a laser rangefinder
($\pm1$~cm). As Table~\ref{tab:depth} shows, empirical calibration reduces the
mean relative error from 27.4\% (raw metric DA2) to below 10\% for both profiles.
This precision is sufficient for all control decisions, which use distance
\emph{intervals} (e.g.\ 1.7--6.0~m for door thresholds, 0.7--1.2~m for the ascent
braking zone) rather than precise absolute values (Fig.~\ref{fig:depthmap}).

\begin{table}[t]
\centering
\caption{Calibrated depth-map error versus ground truth.}
\label{tab:depth}
\footnotesize
\begin{tabular}{@{}lccc@{}}
\toprule
\textbf{Profile} & \textbf{MAE} & \textbf{Mean rel.\ error} & \textbf{Std} \\
\midrule
\texttt{room} (0.5--3.0~m)         & $\sim$0.18~m & 8.5\%  & 0.12~m \\
\texttt{corridor} (1.0--6.0~m)     & $\sim$0.31~m & 9.2\%  & 0.24~m \\
Uncalibrated (raw metric DA2)      & $\sim$0.85~m & 27.4\% & 0.61~m \\
\bottomrule
\end{tabular}
\end{table}

\subsection{Corridor navigation}
\subsubsection{Door detection}
The algorithm was evaluated offline on 142 positive and 168 negative frames
extracted from 8 recordings. With all filters active it reaches a precision of
0.93, a recall of 0.90, and an F1-score of 0.91 (Table~\ref{tab:door}). The
vertical-edge filter is by far the most effective at suppressing false positives,
cutting them from 47 to 9---confirming the core hypothesis that a real door is
characterized more by its vertical casing edges than by the mere presence of a
distant region.

\begin{table}[t]
\centering
\caption{Door-detection filter ablation (offline validation set).}
\label{tab:door}
\footnotesize
\begin{tabular}{@{}lcccc@{}}
\toprule
\textbf{Metric} & \textbf{All filters} & \textbf{No edge} & \textbf{No content} & \textbf{Contour only} \\
\midrule
True positives  & 128 & 134 & 130 & 138 \\
False positives & \textbf{9} & 47 & 28 & 91 \\
Precision       & \textbf{0.93} & 0.74 & 0.82 & 0.60 \\
Recall          & 0.90 & 0.94 & 0.92 & 0.97 \\
F1              & \textbf{0.91} & 0.83 & 0.86 & 0.74 \\
\bottomrule
\end{tabular}
\end{table}

\begin{figure*}[t]
\centering
\begin{subfigure}[t]{0.32\textwidth}\centering
\includegraphics[width=\linewidth]{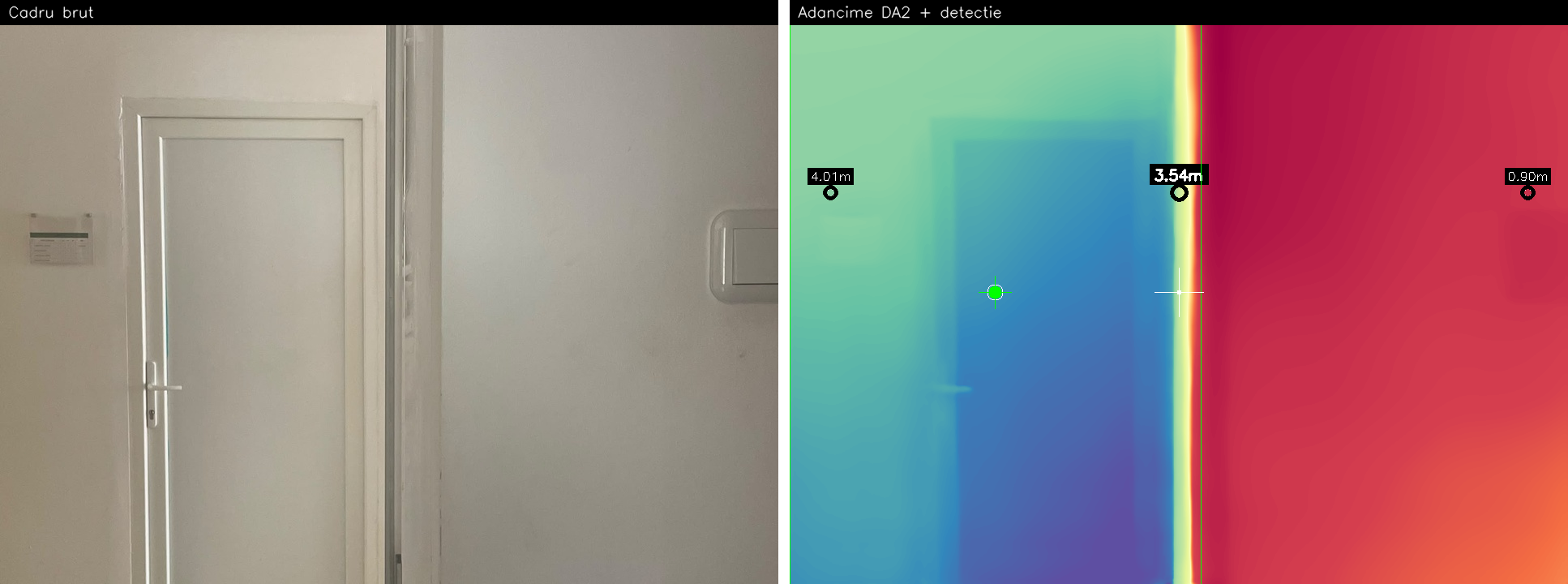}\caption{Partial door, correctly detected ($R=52.75$)}\end{subfigure}\hfill
\begin{subfigure}[t]{0.32\textwidth}\centering
\includegraphics[width=\linewidth]{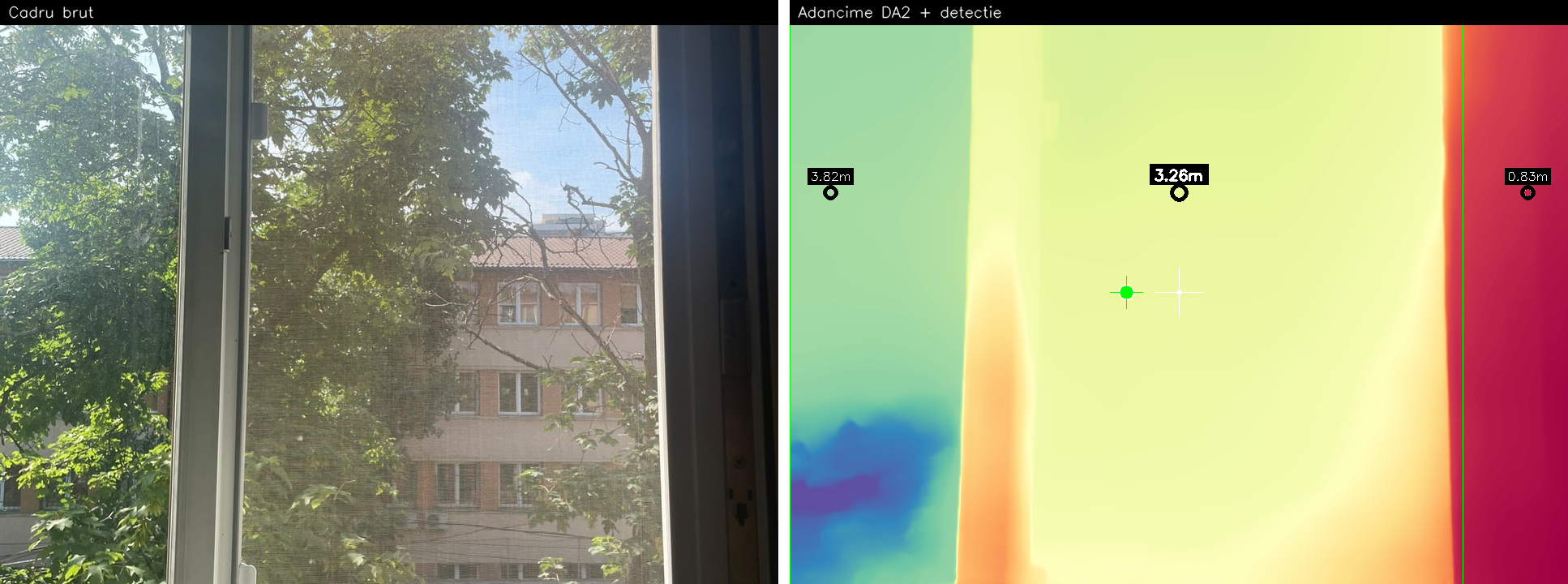}\caption{Window, false positive ($R=6.97$)}\end{subfigure}\hfill
\begin{subfigure}[t]{0.32\textwidth}\centering
\includegraphics[width=\linewidth]{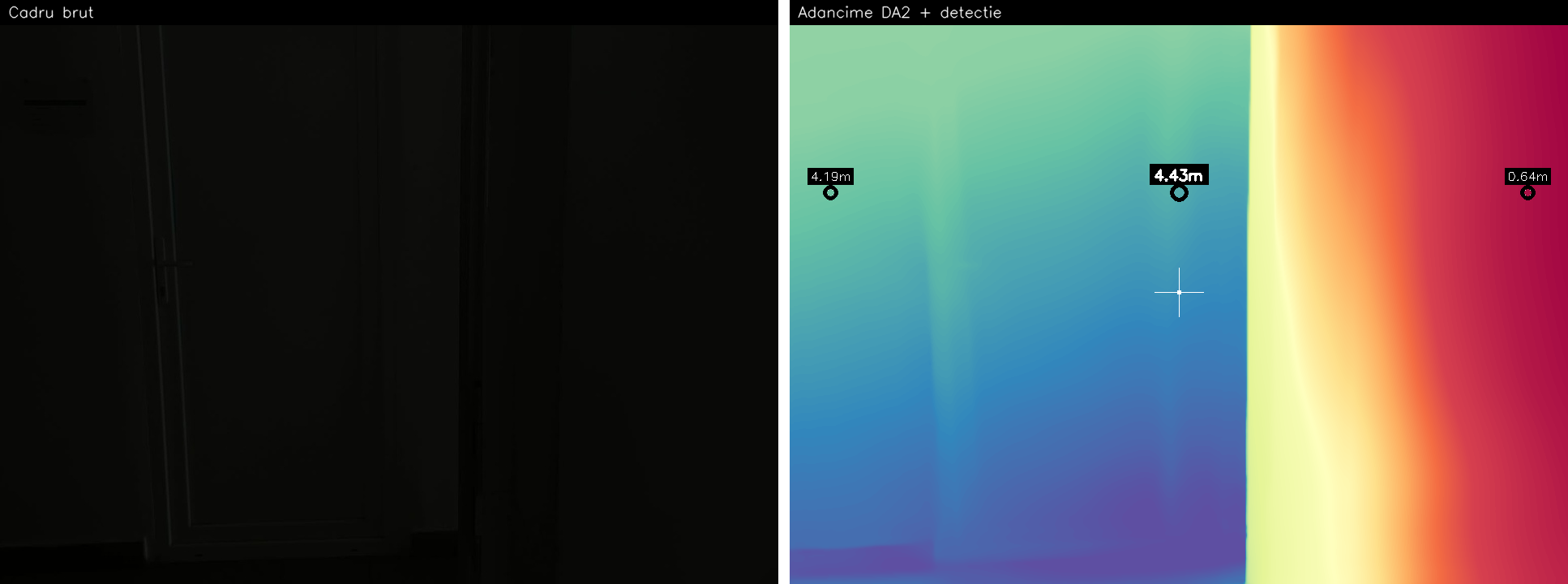}\caption{Low light, false negative ($R=0.10$)}\end{subfigure}
\caption{Door-detection failure analysis. The edge ratio $R$ (threshold $1.20$)
decides acceptance. Appearance traps such as wardrobes and mirrors are rejected;
the two residual failures---glass and very low light---are inherent to monocular
depth, not to the detection logic.}
\label{fig:doorfail}
\end{figure*}

To characterize its limits, the real detector was run on photographs of difficult
scenes, brought to the drone frame format. Two behaviors emerge
(Fig.~\ref{fig:doorfail}). First, \emph{robustness to appearance traps}: a
wardrobe and a mirror both produce a distant region in the depth map, but without
a real casing---their margins are no closer than their centers, so $R$ stays far
below threshold ($0.05$ and $0.38$) and both are correctly rejected. The
detection thus rests on the geometric structure of the opening, not on
appearance. Second, \emph{genuine limitations}: a glass window
(Fig.~\ref{fig:doorfail}b) has exactly the depth profile of an opening, with
clear lateral edges ($R=6.97$), which a purely monocular method cannot
distinguish from a door; and under very low light
(Fig.~\ref{fig:doorfail}c) the depth map becomes too smooth, the real casing
edges vanish ($R=0.10$), and a real door is missed. Both failures stem from the
monocular depth sensor, not from the detection logic---and in operation the
presence of a real door is later confirmed by OCR of the plate on entry.

\subsubsection{Closed-loop behavior}
\begin{figure}[t]
\centering
\includegraphics[width=0.92\columnwidth]{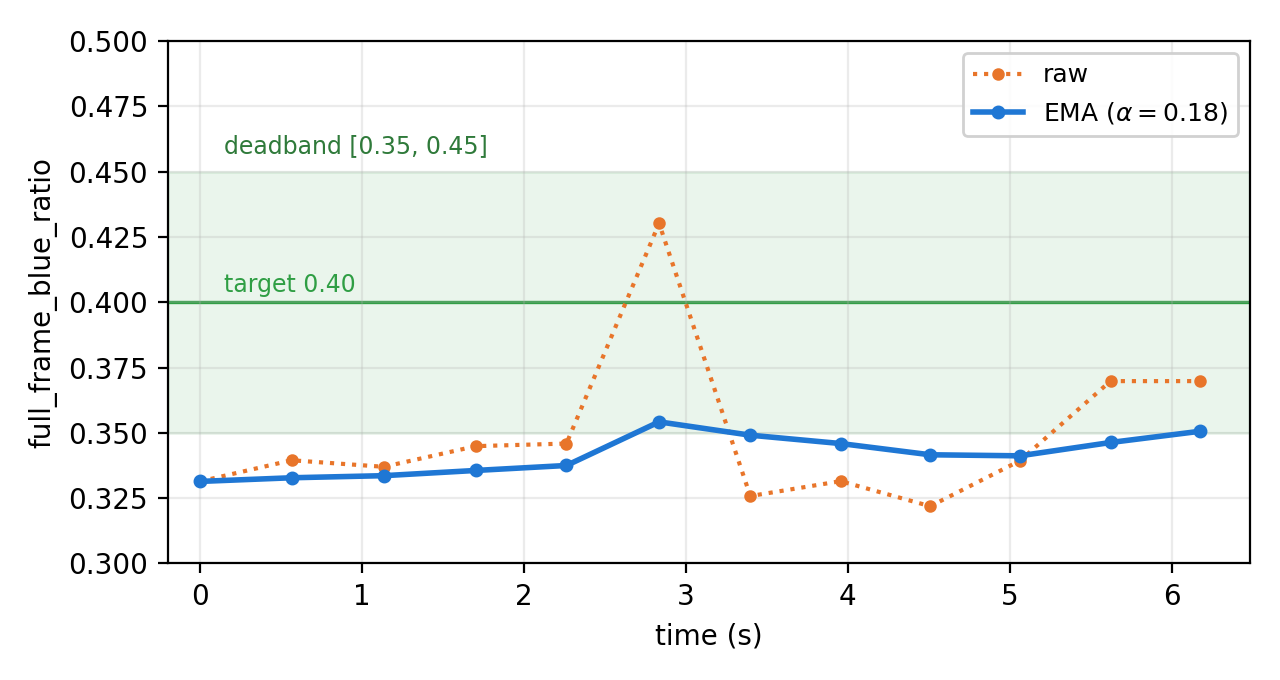}
\caption{Real evolution of \texttt{full\_frame\_blue\_ratio} (raw vs.\ EMA)
during a door entry. The EMA absorbs the raw-signal noise spike and converges
within the deadband, preventing a spurious direction reversal.}
\label{fig:emareal}
\end{figure}

Across 12 end-to-end flights, lateral centering finished within the $\pm25$~px
tolerance in 11 of 12 cases (the single miss still entered without collision).
The distance controller uses \texttt{full\_frame\_blue\_ratio} filtered by the
EMA ($\alpha=0.18$), with a target and a $\pm0.05$ deadband, declaring
stabilization after three consecutive in-band frames.
Fig.~\ref{fig:emareal} shows a real door entry extracted from the mission logs: a
noise spike in the raw signal is absorbed by the EMA, which stays smooth and
converges within the deadband without reversing the drone's direction. Across the
seven logged door entries, the EMA variation stayed small (median $\sim$0.03),
confirming the stability of the filtering; the single large excursion ($0.29$)
corresponds to an entry started from a large distance. Room entry (1.8~m advance)
succeeded in 12/12 cases without touching the door frame. The Simple scan
succeeded in 12/12 flights ($\sim$42~s, 1.6 detected people on average); the
Complex scan succeeded in 10/12 ($\sim$60~s, 3.2 people), the two failures caused
by cumulative yaw drift after two consecutive $90^\circ$ rotations---a known
Tello limitation addressed in future work through IMU gravity-vector
compensation. All 12 missions produced a valid report, OCR read the room number
correctly in 82\% of automatic in-flight attempts, and every effectively visible
person was qualified and reported.

\subsection{Stair ascent: the ablation study}
\label{sec:ablation}
The contribution of each virtual sensor was assessed in two stages: an
\emph{offline} evaluation of detection quality and a \emph{closed-loop} study on
real flights with each signal isolated and each combination active.

\textbf{Offline detection.} On 220 frames labeled \emph{stairs visible} (118),
\emph{landing} (54), and \emph{other} (48), Table~\ref{tab:sensors} reports the
per-sensor and fused precision/recall. Full fusion clearly dominates, especially
in reducing false positives: Sobel false positives on walls with doors
(interpreted as horizontal lines) are removed by Gabor, while Gabor false
positives on textiles or tiled floors are removed by the low-CoV DA2 signal on
flat regions. TRISTAR reaches a precision of 0.92 and a recall of 0.88 on the
\emph{stairs} class, and 0.86/0.81 on the \emph{landing} class---higher than any
single sensor or pair.

\begin{table}[t]
\centering
\caption{Individual virtual sensors versus the fused TRISTAR score, on the
\emph{stairs} and \emph{landing} classes.}
\label{tab:sensors}
\footnotesize
\setlength{\tabcolsep}{4pt}
\begin{tabular}{@{}lcccc@{}}
\toprule
 & \multicolumn{2}{c}{\textbf{Stairs}} & \multicolumn{2}{c}{\textbf{Landing}} \\
\cmidrule(lr){2-3}\cmidrule(lr){4-5}
\textbf{Sensor} & Precision & Recall & Precision & Recall \\
\midrule
Sobel only          & 0.78 & 0.85 & 0.65 & 0.59 \\
Gabor only          & 0.82 & 0.79 & 0.71 & 0.68 \\
DA2 only            & 0.74 & 0.81 & 0.83 & 0.77 \\
Sobel + Gabor       & 0.87 & 0.86 & 0.69 & 0.63 \\
Sobel + DA2         & 0.83 & 0.86 & 0.82 & 0.76 \\
Gabor + DA2         & 0.85 & 0.84 & 0.83 & 0.78 \\
\textbf{TRISTAR}    & \textbf{0.92} & \textbf{0.88} & \textbf{0.86} & \textbf{0.81} \\
\bottomrule
\end{tabular}
\end{table}

\begin{figure*}[t]
\centering
\begin{subfigure}[t]{0.48\textwidth}\centering
\includegraphics[width=\linewidth]{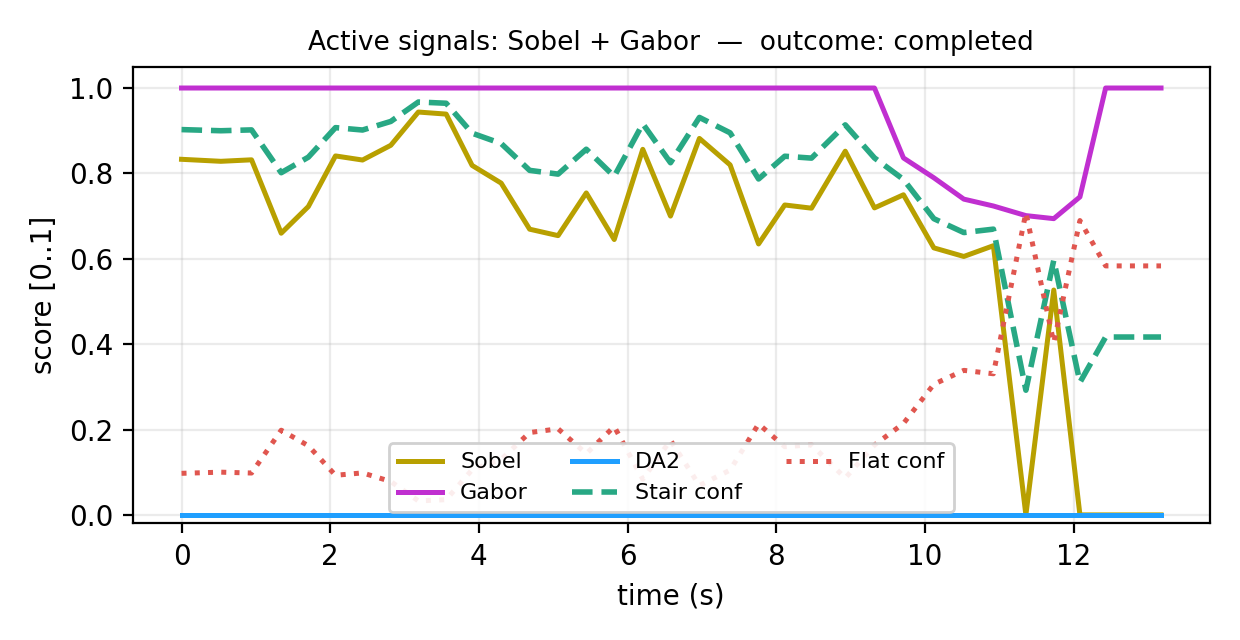}\caption{Sobel + Gabor}\end{subfigure}\hfill
\begin{subfigure}[t]{0.48\textwidth}\centering
\includegraphics[width=\linewidth]{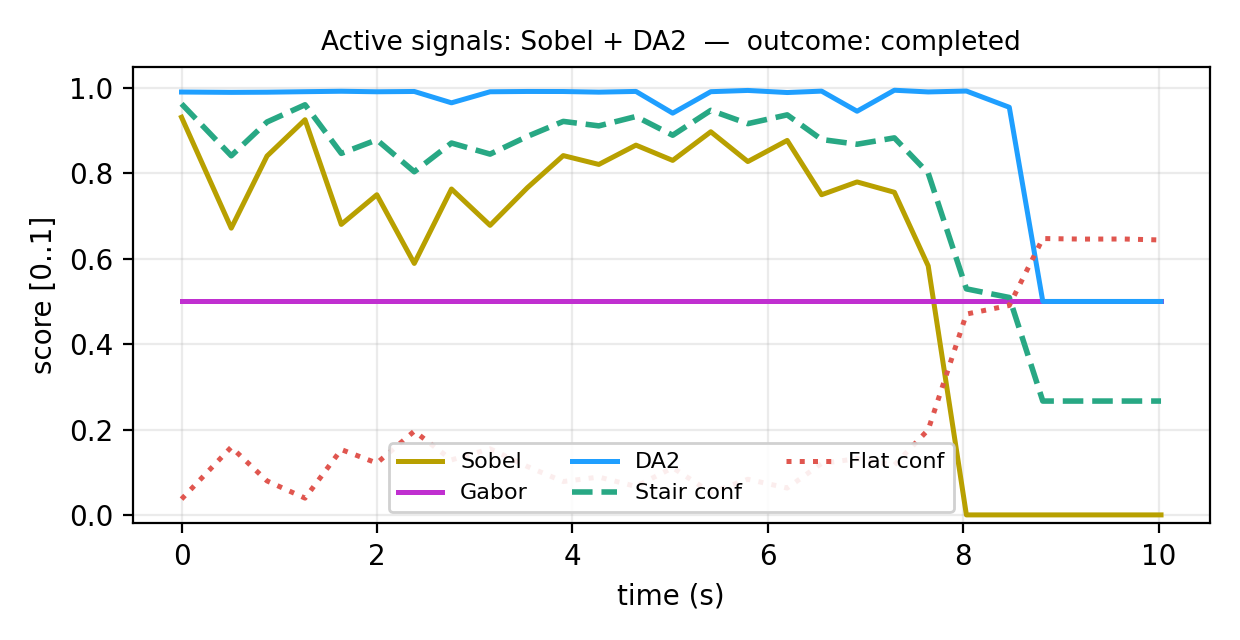}\caption{Sobel + DA2}\end{subfigure}
\\[0.4em]
\begin{subfigure}[t]{0.48\textwidth}\centering
\includegraphics[width=\linewidth]{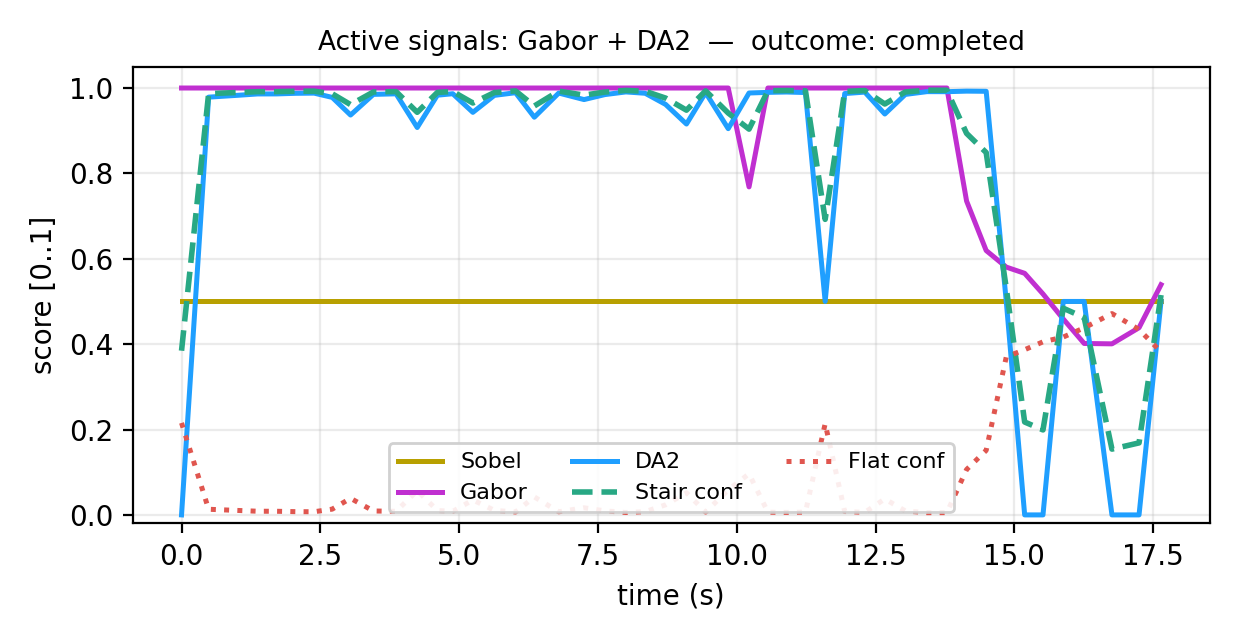}\caption{Gabor + DA2}\end{subfigure}\hfill
\begin{subfigure}[t]{0.48\textwidth}\centering
\includegraphics[width=\linewidth]{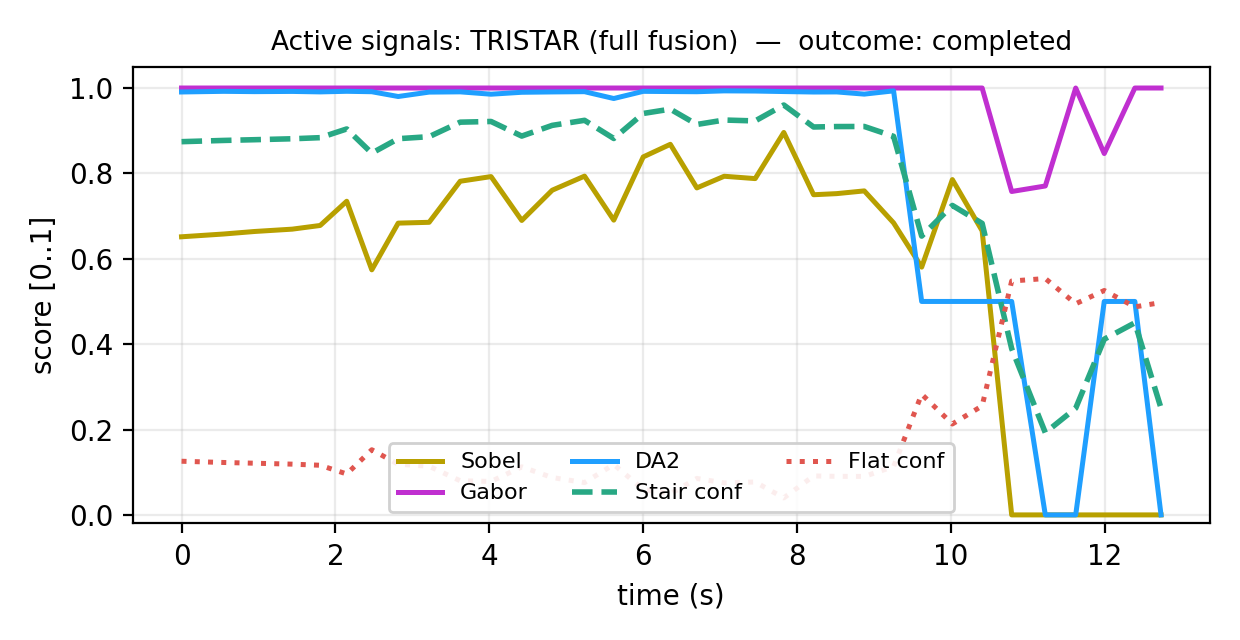}\caption{\textbf{TRISTAR (full fusion)}}\end{subfigure}
\caption{Temporal evolution of the signals during a representative ascent for the
three pairwise configurations and for full fusion. In the pairs, a missing
modality lets one signal drift; under full TRISTAR fusion (d) all signals remain
high and concordant during the climb and jointly collapse (with the ``flat''
confidence rising) exactly at the landing.}
\label{fig:abl}
\end{figure*}

\textbf{Closed-loop flights.} To observe the effect of each signal on in-loop
behavior, the detector was run in seven configurations---each signal alone, the
three pairs, and full fusion---with four ascent flights per configuration
(28 flights total). Table~\ref{tab:ablation} summarizes the outcomes and the
dominant failure mode of each partial configuration. The result is
unambiguous: no partial configuration matches the reliability of full fusion, and
each missing signal induces a characteristic failure. Without depth (Sobel,
Gabor, Sobel+Gabor) the drone is ``blind'' to the end of the ramp and either
overshoots into the ceiling/wall or lands prematurely; DA2 alone is fooled by
window sills that mimic a stair depth profile and never succeeds; removing Sobel
(Gabor+DA2) drops the per-step ``up'' cue and the drone advances past the
landing. Only TRISTAR succeeds in all four flights, each signal covering the
others' blind spots. Figs.~\ref{fig:abl} show, for the three pairs and for full
fusion, the temporal evolution of the signals during a representative flight.

\begin{table}[t]
\centering
\caption{Closed-loop ablation on real flights (4 flights per configuration).}
\label{tab:ablation}
\footnotesize
\begin{tabularx}{\columnwidth}{@{}lcL@{}}
\toprule
\textbf{Configuration} & \textbf{Success} & \textbf{Dominant failure mode} \\
\midrule
Sobel          & 2/4 & Reads steps but, lacking depth, overshoots the landing or lands early. \\
Gabor          & 2/4 & Recognizes the repetitive structure but rises excessively past the last step. \\
DA2            & 0/4 & Depth monotonicity fooled by windows/sills; landing not recognized. \\
Sobel + Gabor  & 1/4 & Strong ``up'' cue but no depth $\Rightarrow$ climbs toward the ceiling. \\
Sobel + DA2    & 2/4 & Works but unstable: useless lifts and a false landing when steps leave view. \\
Gabor + DA2    & 1/4 & No per-step ``up'' cue $\Rightarrow$ too many forward commands, overshoots. \\
\textbf{TRISTAR} & \textbf{4/4} & \textbf{Full fusion: every signal covers the others' weaknesses.} \\
\bottomrule
\end{tabularx}
\end{table}

\textbf{Failure-case analysis.} Running the real detector on static photographs
of deceptive scenes (Table~\ref{tab:failcases}) confirms the role of fusion. On a
railing and on a shelf, Sobel and Gabor respond strongly to the repetitive
horizontal lines (a per-signal false positive), but the DA2 component ($0.00$)
pulls the composite score below threshold, correctly rejecting the scene
(the \emph{Railing} row of Table~\ref{tab:failcases}). Conversely, the system
misses a scene (false negative)
when the steps are not geometrically visible: an obstacle in front of the
staircase or a field of view with too few steps simultaneously nullifies Sobel
and DA2, and Gabor alone cannot reach the threshold. These modes confirm that
detection depends on the geometric visibility of the steps, while robustness to
deceptive structures is provided precisely by the fusion of the three signals,
not by any single detector.

\begin{table}[t]
\centering
\caption{Per-signal scores and the TRISTAR decision (ascent if $\ge0.52$).}
\label{tab:failcases}
\footnotesize
\begin{tabular}{@{}lccccc c@{}}
\toprule
\textbf{Scene} & \textbf{Truth} & \textbf{Sobel} & \textbf{Gabor} & \textbf{DA2} & \textbf{Score} & \textbf{OK?} \\
\midrule
Stairs, low light      & stairs & 0.80 & 1.00 & 0.99 & 0.93 & \yes \\
Partial stairs (left)  & stairs & 0.75 & 1.00 & 1.00 & 0.91 & \yes \\
Stairs + obstacle      & stairs & 0.00 & 1.00 & 0.00 & 0.25 & \no  \\
Partial stairs (right) & stairs & 0.00 & 1.00 & 0.00 & 0.25 & \no  \\
Railing                & non    & 0.68 & 1.00 & 0.00 & 0.49 & \yes \\
Shelf                  & non    & 0.85 & 0.69 & 0.00 & 0.47 & \yes \\
\bottomrule
\end{tabular}
\end{table}

For a typical ramp (12 steps, $\sim$3.4~m long, $\sim$2.05~m high), the drone
climbed in $\sim$19~s on average, at an effective vertical speed of
$\sim$0.11~m/s---below the drone's theoretical maxima, reflecting the
conservative control policy (braking near the steps, five-frame confirmation for
landing detection).

\subsection{Auxiliary AI components}
Posture estimation from YOLO11-Pose keypoints, evaluated on 96 frames, reached a
weighted accuracy of $\sim$87\%, with the operationally critical \emph{fallen}
class at 83\%. OCR of room numbers reached 85\% overall (92\% on
favorable-angle frames, 82\% on automatic in-flight frames), with failures traced
to strong backlight, capture angle, and non-standard plate fonts. The Gemini
medical triage returned an acceptable state (STABLE/UNKNOWN) in 22 of 24 cases,
with the two over-classifications (INJURED on a slightly tilted posture)
illustrating the tendency of large models to over-interpret subtle visual cues;
the mean response time was $\sim$1.6~s, small enough to be included in the report
without delaying post-processing. The fire/smoke detector reached 8/8 true
positives and 1/12 false positives on reference images and runs as a separate
reporting thread that does not affect navigation.

\subsection{Summary}
Table~\ref{tab:summary} consolidates the main results against their implicit
targets. All targets are met or exceeded, and the two autonomous subsystems
operate independently on real flights. The identified limitations---cumulative
yaw drift over multiple rotations, dependence on Gemini for medical triage, and
the absence of a global map---are known, documented, and form the direct agenda
of future iterations.

\begin{table}[t]
\centering
\caption{Summary of results against implicit targets.}
\label{tab:summary}
\footnotesize
\begin{tabularx}{\columnwidth}{@{}Lccc@{}}
\toprule
\textbf{Component / metric} & \textbf{Target} & \textbf{Achieved} & \\
\midrule
Raw stream FPS                 & $\ge25$   & $\sim$30       & \yes \\
Depth stream FPS               & $\ge10$   & $\sim$10--12   & \yes \\
End-to-end latency             & $<300$~ms & $\sim$85--100~ms & \yes \\
Calibrated depth error (room)  & $<15\%$   & $\sim$8.5\%    & \yes \\
Door detection F1 (offline)    & $>0.85$   & 0.91           & \yes \\
Corridor flight success        & $>80\%$   & 12/12          & \yes \\
Stair detection F1 (offline)   & $>0.85$   & $\sim$0.90     & \yes \\
Stair ascent success (flight~1)& $>80\%$   & 8/8            & \yes \\
Room-number OCR accuracy       & $>80\%$   & 85\%           & \yes \\
Posture accuracy               & $>75\%$   & $\sim$87\%     & \yes \\
\bottomrule
\end{tabularx}
\end{table}

\section{Conclusion}
\label{sec:conclusion}
This paper set out from a deliberately practical question---\emph{how much can be
done autonomously with a cheap commercial drone that carries no dedicated
sensors}---and answered it concretely. We showed that modern monocular vision,
combined with classical computer-vision tools and a clean modular design, is
enough to let a sub-150~EUR drone autonomously explore a building corridor, enter
rooms, read their labels, detect and triage people, and climb a staircase---
tasks that until now demanded either a human operator or an expensive sensor
suite. The central technical result is that the division of labor between
paradigms pays off: geometric structure (open doors, stair steps) is recovered
elegantly from a calibrated monocular depth map and classical filters, without
training a dedicated detector, while semantic understanding (people, fire, smoke,
medical state) is delegated to neural models. Empirical calibration alone brings
the relative depth error from 27.4\% to below 10\%, enough for every control
decision in the system, and the door detector reaches an F1-score of 0.91 while
correctly rejecting appearance traps such as wardrobes and mirrors.

The clearest embodiment of this philosophy is TRISTAR. The ablation study is
categorical: no single virtual sensor, and no pair of them, matches the
reliability of the three-way fusion---each isolated modality carries a
characteristic, reproducible failure mode, and only the composite score succeeds
in every flight. Structural cues supply the per-step signal, texture analysis
supplies robustness to perspective, and monocular depth supplies the geometric
validation that recognizes the landing and refuses the traps that fool the other
two. It is precisely because each sensor is individually fragile that their
fusion is strong.

Two honest limitations frame these results. The system builds no persistent
(SLAM) map, so each mission is geometrically independent, and the corridor and
stair behaviors are still operated separately rather than composed into a single
continuous mission. Neither limitation is fundamental: both follow largely from
the chosen hardware, not from the software stack, which is in principle portable
to any drone exposing a video feed and RC control. The most promising directions
are therefore a monocular-SLAM layer for a persistent building map, a transition
state that chains stair ascent into upper-floor corridor exploration, on-device
medical triage to remove the cloud dependency, and a migration to a more capable
platform for multi-drone coordinated exploration.

More broadly, this work illustrates a wider trend in applied robotics: as vision
models grow more capable and more efficient, the total cost of a drone capable of
genuine indoor autonomy keeps falling, and the accessibility of such systems
keeps rising. A capability that today requires an 11{,}000~USD platform is,
increasingly, a matter of software running on an 80-gram toy. If a commodity
camera and a few classical filters can already guide a drone up a dark stairwell
toward the people who need to be found, the question worth carrying forward is no
longer whether low-cost autonomy is possible, but how far it can be pushed---and
how many lives that reach might one day help save.


\end{document}